\documentclass[conf]{new-aiaa}
\usepackage[utf8]{inputenc}

\usepackage{graphicx}
\usepackage{amsmath}
\usepackage[version=4]{mhchem}
\usepackage{siunitx}
\usepackage{longtable,tabularx}
\title{Data-driven Analysis of Turbulent Flame Images}

\author{Rathziel Roncancio\footnote{Graduate Research Assistant, AIAA Member.}}
\affil{Purdue University, West Lafayette, IN 47907, USA}
\author{Jupyoung Kim\footnote{Researcher, Aerospace R\&D center, Hanwha Aerospace, South Korea and AIAA Member.}}
\affil{Aerospace R\&D center, Hanwha Aerospace, South Korea}
\author{Aly El Gamal\footnote{Professor, School of Electrical and Computer Engineering}, and Jay P. Gore\footnote{Reilly University Chair Professor, School of Mechanical Engineering, AIAA Fellow}}
\affil{Purdue University, West Lafayette, IN 47907, USA}

\begin{document}

\maketitle

\begin{abstract}
Turbulent premixed flames are important for power generation using gas turbines. Improvements in characterization and understanding of turbulent flames continue particularly for transient events like ignition and extinction. Pockets or islands of unburned material are features of turbulent flames during these events. These features are directly linked to heat release rates and hydrocarbons emissions. Unburned material pockets in turbulent CH$_4$/air premixed flames with CO$_2$ addition were investigated using OH Planar Laser-Induced Fluorescence images. Convolutional Neural Networks (CNN) were used to classify images containing unburned pockets for three turbulent flames with 0\%, 5\%, and 10\% CO$_2$ addition. The CNN model was constructed using three convolutional layers and two fully connected layers using dropout and weight decay. The CNN model achieved accuracies of 91.72\%, 89.35\% and 85.80\% for the three flames, respectively. \end{abstract}

\section{Introduction}

Turbulent flows are widely used in gas turbines because of increases in fuel air mixing and associated high heat release rates in the combustion process.  Gas turbines are used in aircraft and stationary power systems. Their use is increasing because of increases in air travel and in rapid deployment of backup power for renewable energy sources such as wind and solar.  Challenges associated with turbine combustion include the simultaneous control of soot, carbon monoxide, and emissions of nitrogen oxides \cite{CORREA1993,Turns2000}. Great efforts are invested in reducing the nitrogen oxides' emissions \cite{tanaka2013development} ranging from novel injector designs to stage combustion and beyond \cite{Lefebvre1995}. Emissions from gas turbine engines impact air quality through the change in the chemistry of the upper troposphere and stratosphere. However, the full environmental impact of emissions from gas turbines remains a matter of study. With the advent of new alternative fuels to meet the world demand and growing environmental concerns, a new series of complex hydrocarbon emissions need to be studied \cite{Zeppieri2014}.

One of the prominent characteristic features of turbulent flames near extinction is the formation of unburned reactant pockets \cite{poinsot1991diagrams}. These pockets can result in excessive carbon monoxide and unburned hydrocarbons emissions.  The prevalence of the pockets increases with high turbulence and high exhaust gas recirculation (EGR) \cite{Roberts1991}. Enhanced turbulent burning velocity leads to a thickening of the flame brush and rapid consumption of unburned reactant pockets near the flame tip \cite{johchi2015investigation}. Pocket formation and consumption processes have substantial implications in the development of accurate Large Eddy Simulation (LES) models based on their size in relation to the turbulence length scales including the Taylor microscale and the integral scale \cite{Tyagi2020, Kim2019}. Unburned pocket formation and consumption at the tip of a Bunsen flame \cite{Filatyev2005,Kobayashi2005}, and turbulent jet premixed flames \cite{Li2010, johchi2015investigation, Han2018} have been observed. Characterization of unburned pockets in turbulent premixed flames is of importance in the understanding of flame burning velocity and control of pollutant emissions.

Artificial intelligence is an emerging field with numerous practical applications. Machine learning is one of the approaches to artificial intelligence involving extraction of knowledge in the form of patterns within the data. This allows computers to suggest subjective decisions. Examples of machine learning algorithms are naive Bayes, support vector machines, and logistic regression. By drawing information from past operational data, these algorithms suggest adjustment of operating parameters aimed at achieving object functions while continuously generating the scientific rules of the systems in a heuristic manner. The patterns extracted from the data build onto each other creating stacked layers. These layers of rules continue to increase the depth of the stack that builds leading to the name \textit{Deep Learning (DL)} to the approach \cite{Goodfellow-et-al-2016}. Programming frameworks with high levels of abstraction have been made available in the past four years, in conjunction with powerful hardware such as GPUs to perform fast training \cite{Abadi2016}. These developments have opened the possibility for applications in many fields, such as combustion, where the causal nature of DL suggests that intricate patterns could also be sought and learned.

Lapeyre et al. \cite{Lapeyre2019} formulated the sub-grid surface density estimation problem using machine learning. A Convolutional Neural Network (CNN) was trained on the results of a Direct Numerical Simulation (DNS) of a premixed turbulent flame. The CNN efficiently extracted the flame topology and estimated the subscale wrinkling. The CNN outperformed two algebraic models, one based on a power-law expression and another based on fractal surfaces. Barwey et al. \cite{Barwey2019} used a CNN to obtain three-component velocity fields from a series of time-resolved OH-PLIF images in a premixed swirl combustor. Two types of models were compared, a global CNN trained on a 472 px $\times$ 408 px domain and a set of local CNNs, trained on individual 118 px $\times$ 136 px sections of the domain. They tested the models on two different sets, one having the same operating conditions as the training set and another having different operating conditions. The data sets contained images in both attached and detached flame regimes. They found that for both sets, both CNN models performed considerably better in the attached flame regime.

In the present study, OH-PLIF images of Han et al. \cite{Han2018} are analyzed using a convolutional neural network (CNN) model. The model is applied to the classification of thousands of images of three flames with CO$_{2}$ dilutions of 0\%, 5\%, and 10\% with some images containing unburned pockets represented by a lack of the OH-PLIF signal. The results show that CNN models can be very effective in analyses of large data sets resulting from quantitative imaging of turbulent combustion.

\section{Test configuration and Computational Methodology}

\subsection{Burner Details}
\begin{figure}[!h]
\centering
\includegraphics[width=0.6\textwidth]{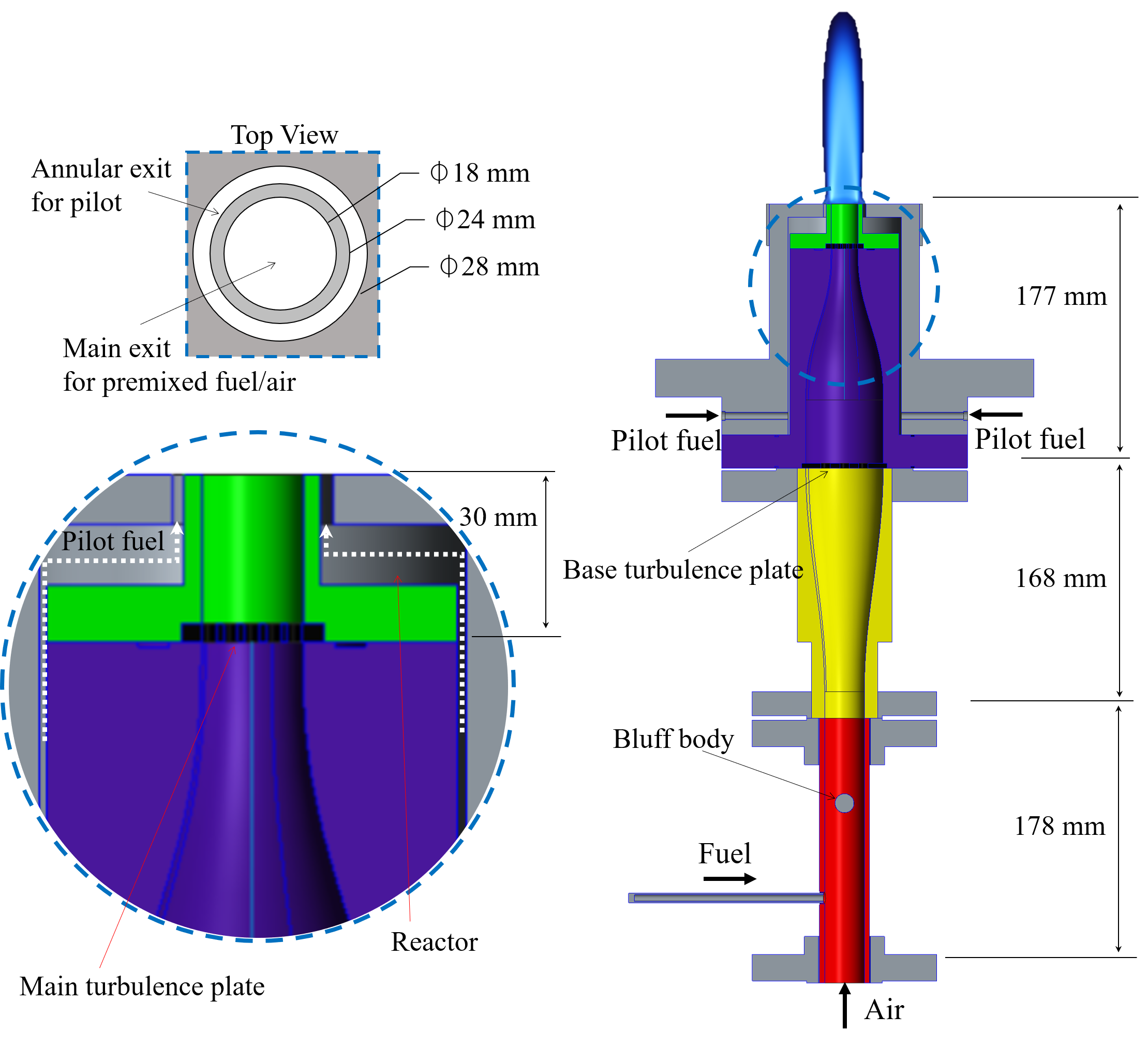}
\caption{A cross-sectional schematic of the piloted axisymmetric reactor assisted turbulent burner (PARAT) burner \cite{Kim2020}.}
\label{Fig:burnerfigure}
\end{figure}

The piloted axisymmetric reactor assisted turbulent burner (PARAT) employed by Han et al. \cite{Han2018} and Kim et al. \cite{Kim2020} to study lean turbulent methane-air premixed flames is shown in Fig. \ref{Fig:burnerfigure}.

The burner consists of four components: a fuel/air injection component (red), a diverging duct component (yellow), a converging duct component (purple), and an exit component (green). The gray component in Fig. \ref{Fig:burnerfigure} is the casing of the burner that provides the frame for the remaining four components. The fuel/air injection component (red) is 178 mm long, the diverging duct component (yellow) is 168 mm long, the converging duct component (blue) is 147 mm long, and the exit component (green) is 30 mm long. The duct diameter varies from 18 mm to 27 mm along the longitudinal direction of the burner. 

A premixed fuel/air stream flows out through the main exit. The pilot fuel hydrogen flows into separately embedded feeding lines and is ejected through an annular exit. The path of the pilot fuel stream is depicted in the Fig. \ref{Fig:burnerfigure} \cite{Kim2020}. Air is admitted to the inlet of the fuel/air injection component and fuel is injected from the side of the fuel/air injection component. The fuel and air mix thoroughly at the molecular level while reaching the burner exit. 

High turbulence results from the steep velocity gradients at the wall caused by the high Reynolds number and the wake flows of the bluff body and the jets emerging from the two perforated plates. Prevention of flow separation and recirculation zones near the duct wall is obtained using diverging and converging cross sections of the burner. Turbulent eddies generated in the duct are further broken down by the turbulent base plate and finally approach fine scale after passing through the main turbulence plate. Further description of the experimental apparatus is provided in Ref. \cite{Kim2020}.

Planar laser-induced fluorescence (PLIF) using OH $Q_{1}(8)$ transition in the $X^{2}\Pi(v^{"}=0) \rightarrow A^{2}\Sigma^{+}(v^{'}=1)$ vibrational band led to 480x 640 px images. The images were acquired at a repetition frequency of 9 kHz. Specific details of the equipment used for the acquisition of the images is provided in Refs. \cite{Han2018} and \cite{Kim2020}.

\subsection{Convolutional Neural Network Model}

Three flames with Re 10,000 were used: flame 1 (f1) has 0\% CO$_{2}$, flame 2 (f2) has 5\% CO$_{2}$, and flame 3 (f3) has 10\% CO$_{2}$ with equivalence ratios of 0.80, 0.84 and 0.89, respectively \cite{Han2018}, were utilized in Refs. \cite{Han2018,Kim2020} and the resulting data are used in the development of the present CNN model.

\begin{figure}[h]
\centering
\includegraphics[width=0.57\textwidth]{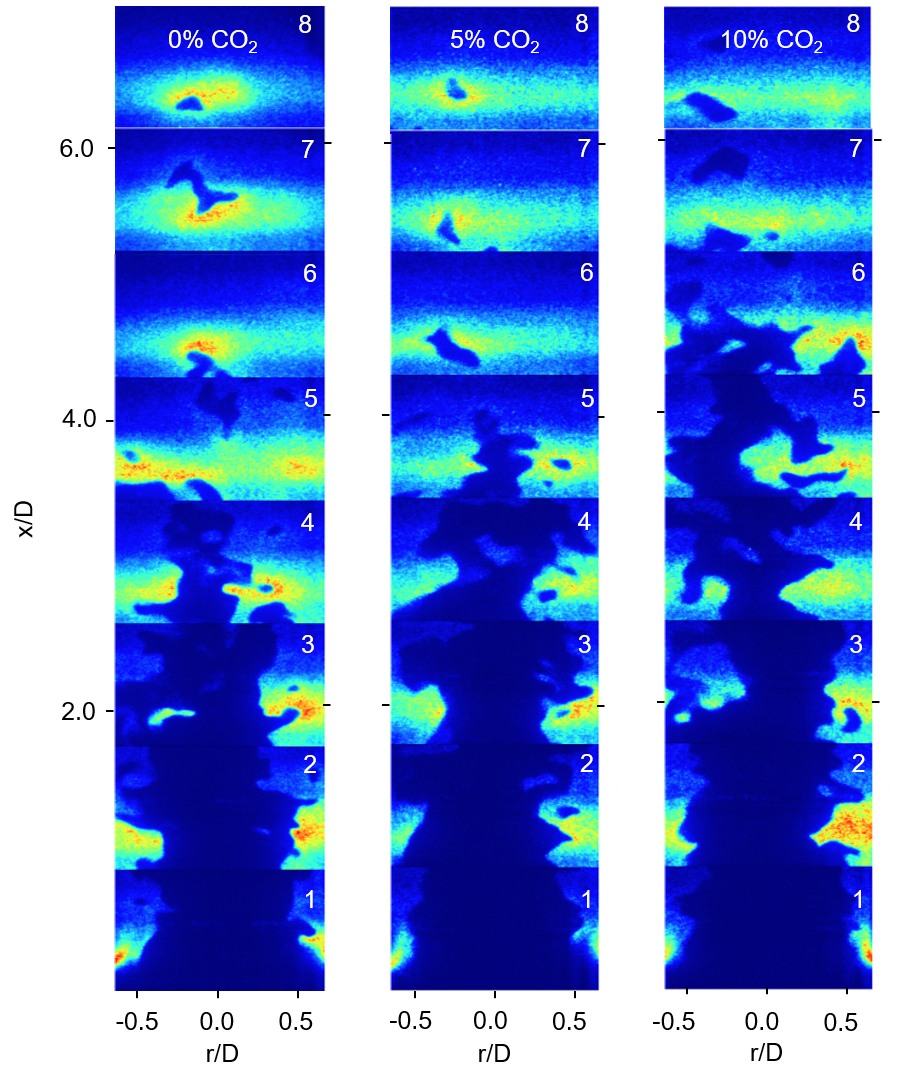}
\caption{Raw instantaneous OH PLIF images used to train the CNN model.}
\label{Fig:Flames}
\end{figure}

Eight panels of OH PLIF images of the three flames, representing 24000 images, are shown in Fig. \ref{Fig:Flames}. The panels are spatially correlated but temporally independent. Flame height increased with the addition of CO$_2$. Pockets are created at the flame brush and travel downstream. The structure of the flames changes beyond the tip where the reactions are completed. Panels four through eight depict the post flame tip region. Panels one through three contain  the main conical bodies of the flames. Outside the conical bodies of the flames, unburned pockets of various sizes appear. The pockets within the length of the conical bodies are smaller than those appearing beyond the tip. These larger pockets break down  further away from the flame tip.  The images were presented to the CNN model during training in batches, each of size 256. A parameter update step is carried out after processing each batch. 
The number of epochs, or complete passes through the training set, used for training the CNN model was 200. 
All of the 24000 images were also examined manually and classified into the two categories to provide a standard for checking the performance of the CNN model. 

\begin{figure}[!ht]
\centering
\includegraphics[width=0.5\textwidth]{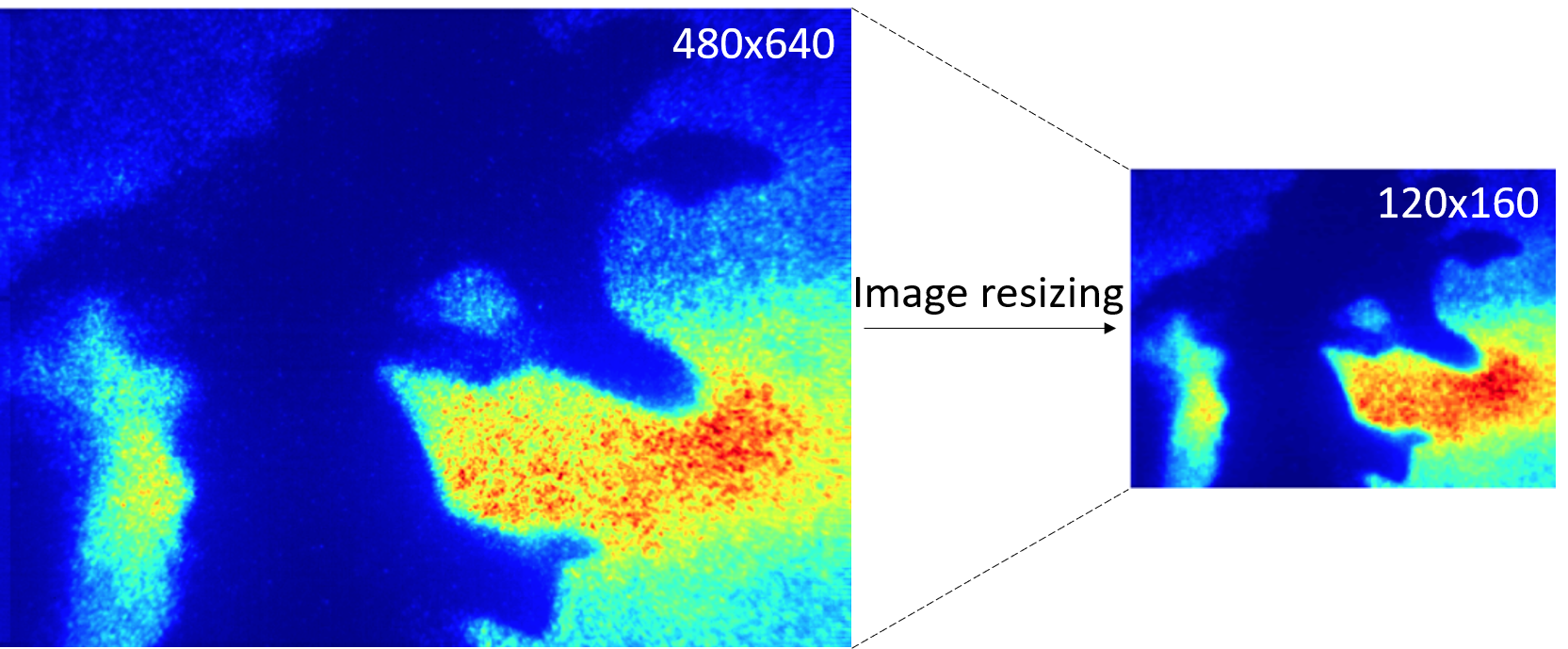}
\caption{Resizing of input images, the images were reduced using bicubic interpolation to maintain sharpness.}
\label{Fig:Resizing}
\end{figure}

The images were resized to 120 px $\times$ 160 px using bicubic interpolation aiming to maintain sharpness of the original image for reducing the computational cost of training the CNN model. Fig. \ref{Fig:Resizing}. illustrates the resizing process. 

\begin{figure}[!ht]
\centering
\includegraphics[width=0.7\textwidth]{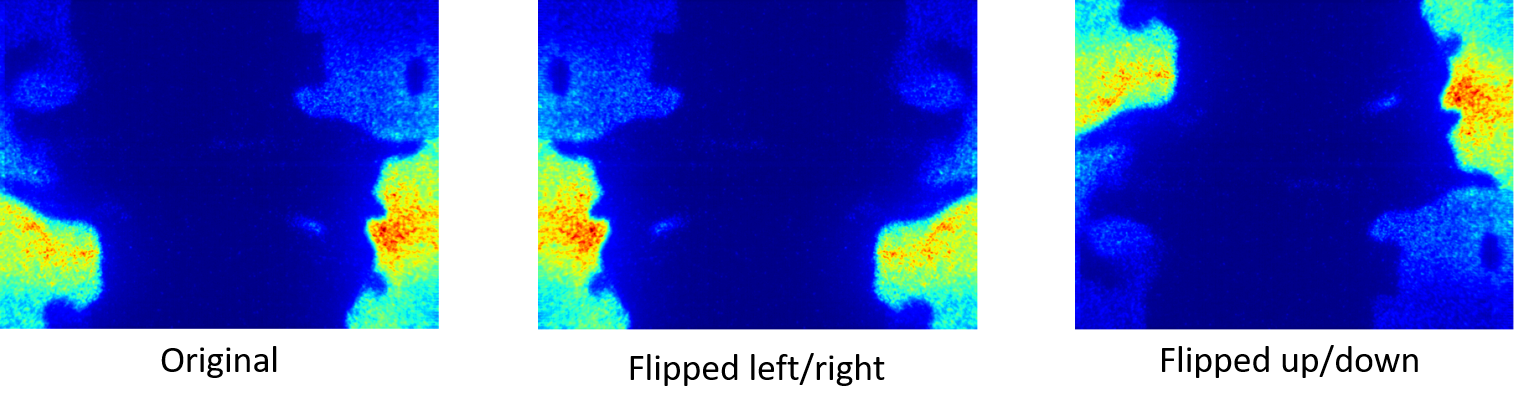}
\caption{Data augmentation techniques used for balancing the data set.}
\label{Fig:Image_Aug}
\end{figure}

Unbalanced classification problems are defined to contain large data sets with much smaller number of events of user interest \cite{Sun2009,Ertekin2007}. The unburned pocket formation problem is by definition an unbalanced problem resulting in unbalanced data. Unbalanced data may be augmented by utilizing user generated data. Specifically, up and down flipping and left and right flipping of actual data may be used for augmentation as shown in Fig. \ref{Fig:Image_Aug}. One of the main tasks of a data-driven model is to remain invariant to input transformations that do not affect the learning task, which typically improves generalization performance \cite{Mikoajczyk2018}. The use of affine transformations, as depicted in Fig. \ref{Fig:Image_Aug}, has been reported to outperform other data augmentation techniques \cite{Kan2018}. The data were divided into three subsets, training (80\%), validation (10\%), and testing (10\%) subsets. We finally note that only the training data set was augmented. 

\begin{figure}[!ht]
\centering
\includegraphics[width=0.96\textwidth]{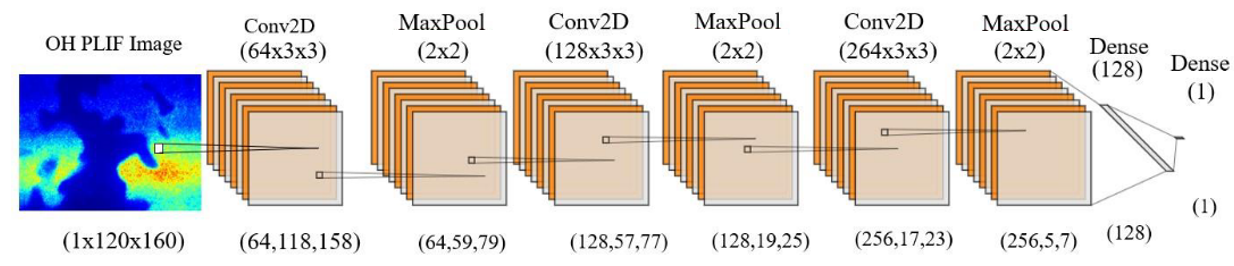}
\caption{Convolutional Neural Network architecture used for the present work. The kernel shape is given below the layer title. The tensor output for each layer is displayed on the lower row.}
\label{Fig:CNN}
\end{figure}

Fig \ref{Fig:CNN} and Table \ref{Table:Layers} depict the structure of the present CNN. The CNN includes 64, 128 and 256 kernels, with respect to order of the hidden layer, with 3$\times$3 filters and followed by a Max Pooling layer with a 2$\times$2 pool size. The CNN is implemented using Keras \cite{chollet2015keras} libraries and TensorFlow backend \cite{tensorflow2015-whitepaper}. $L_2$ regularization is implemented to alleviate overfitting. The weight decay coefficient of $L_2$ regularization is 0.0005 as in \cite{Krizhevsky2017}. After the convolution stage, we flatten the intermediary vector and process the output by two dense layers, with 128 and 1 units, with respect to order. Dropout is applied to the 128-unit dense layer with a unit removal probability of 0.3 (see \cite[Chapter $7$]{Goodfellow-et-al-2016} for more details on Dropout training and weight scaling inference). 

\begin{table}[h]
    \caption{Detailed parameters used for all layers in the CNN model.}
    \label{Table:Layers}
	\centering 
	\resizebox{0.7\columnwidth}{!}{%
	\begin{tabular}{l l l l} 
		\hline\hline 
		Layer & Type & Number of Kernels & Other Layer Parameters\\ [0.5ex] 
		\hline 
		1 & Conv + ReLU & 64 & filter=(3,3), weight decay = 0.0005, Parameters: 640 \\ 
		2 & Max-pooling & 2 & pool\_size=(2,2)\\
		3 & Conv + ReLU & 128 & filter=(3,3), weight decay = 0.0005, Parameters: 73856\\
		4 & Max-pooling & 2 & pool\_size=(2,2)\\
		5 & Conv + ReLU & 256 & filter=(3,3), weight decay = 0.0005, Parameters: 295168\\
		6 & Max-pooling & 2 & pool\_size=(2,2)\\
		7 & Flatten & - & -\\
		8 & FC + ReLU + Dropout & 128 & Dropout\_ratio = 0.3, Parameters: 1147008\\
		9 & FC + Sigmoid & 1 & Parameters: 129\\

		\hline 
	\end{tabular}
	} 

\end{table}

Accuracy, as defined below, is used to gauge the performance of the CNN model \cite{Hao2019}:

\begin{equation}
\label{eq:Accuracy}
Accuracy = \frac{TP+TN}{FP+TN+TP+FN}
\end{equation}

where TN ( true negative) and TP (true positive) are the numbers of images that were correctly classified, and FN (false negative) and FP (false positive) are the numbers of images that were falsely classified. Binary cross-entropy loss is used to measure the performance of the CNN model, and compute the parameter updates during training. 


\begin{equation}
\label{eq:Loss}
J(\theta) = -\mathbb{E}_{x,y\sim \hat{p}_{data}} \log p_{model}(y|x)
\end{equation}

Eq. \ref{eq:Loss} shows the cross-entropy loss function, where \textit{$\theta$} refers to the parameters of the hidden layers, \textit{x} corresponds to the input data and \textit{y} corresponds to the model predicted values. $\mathbb{E}_{x,y\sim\hat{p}_{data}}$ corresponds to the expected value with respect to the empirical distribution of the data ($\hat{p}_{data}$) and $p_{model}$ refers to the model distribution. The loss function penalizes the CNN predicted values that are wrongly classified. The Root Mean Square Propagation (RMSProp) algorithm \cite{tieleman2012lecture} was used as optimizer for the CNN model. A learning rate $\epsilon$ of 0.001 and a decay rate $\rho$ of 0.9 were selected as hyperparameters for the RMSProp algorithm.

\section{Results and Discussion}

\begin{figure}[!ht]
\centering
\includegraphics[width=0.5\textwidth]{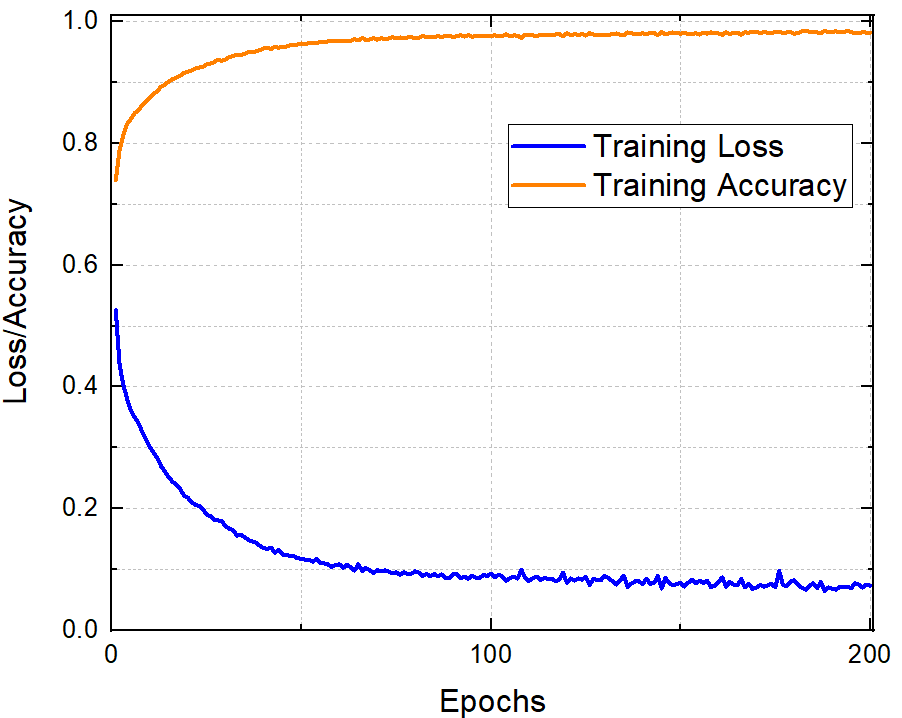}
\caption{Changes in accuracy and loss ratio for the training of the CNN.}
\label{Fig:Accuracy_Loss}
\end{figure}

Fig. \ref{Fig:Accuracy_Loss} shows the accuracy and loss during training as a function of number of epochs. The accuracy increases until it reaches a plateau at 98\%. The loss decreases rapidly for the first 50 epochs. As training takes place, it continues to decrease at a lower rate reaching a final value of 0.07. The loss and accuracy progress as the CNN model learns the features required for the classification task. 

\begin{table}[htbp]
    \caption{CNN model accuracy for each flame}
    \label{Table:Tests}
	\centering 
	\resizebox{0.5\columnwidth}{!}{%
	\begin{tabular}{c c c c c c} 
		\hline\hline 
		Flame & FP & FN & TP & TN & Test Accuracy   \\ [0.4ex] 
		\hline 
		f1 & 201 & 462 & 1037 & 6308 & 91.72\%  \\ 
		f2 & 76 & 777 & 597 & 6558 & 89.35\%   \\
		f3 & 15 & 1122 & 193 & 6678 & 85.80\%   \\
		\hline 
	\end{tabular}
	}
\end{table}

The accuracy of the CNN model on each flame was tested. Table \ref{Table:Tests} displays the accuracy obtained for each flame. The best results were obtained for flame 1, followed by those for flame 2. The number of false negatives increases as the flame height increases while the number of false positives displays an opposite trend. The CNN model performs better for pockets within the conical body of the flame (flame 1), while capturing pockets past the tip of the flame present a limitation for the present model (flame 2 and flame 3).

\begin{figure}[!ht]
\centering
\includegraphics[width=0.9\textwidth]{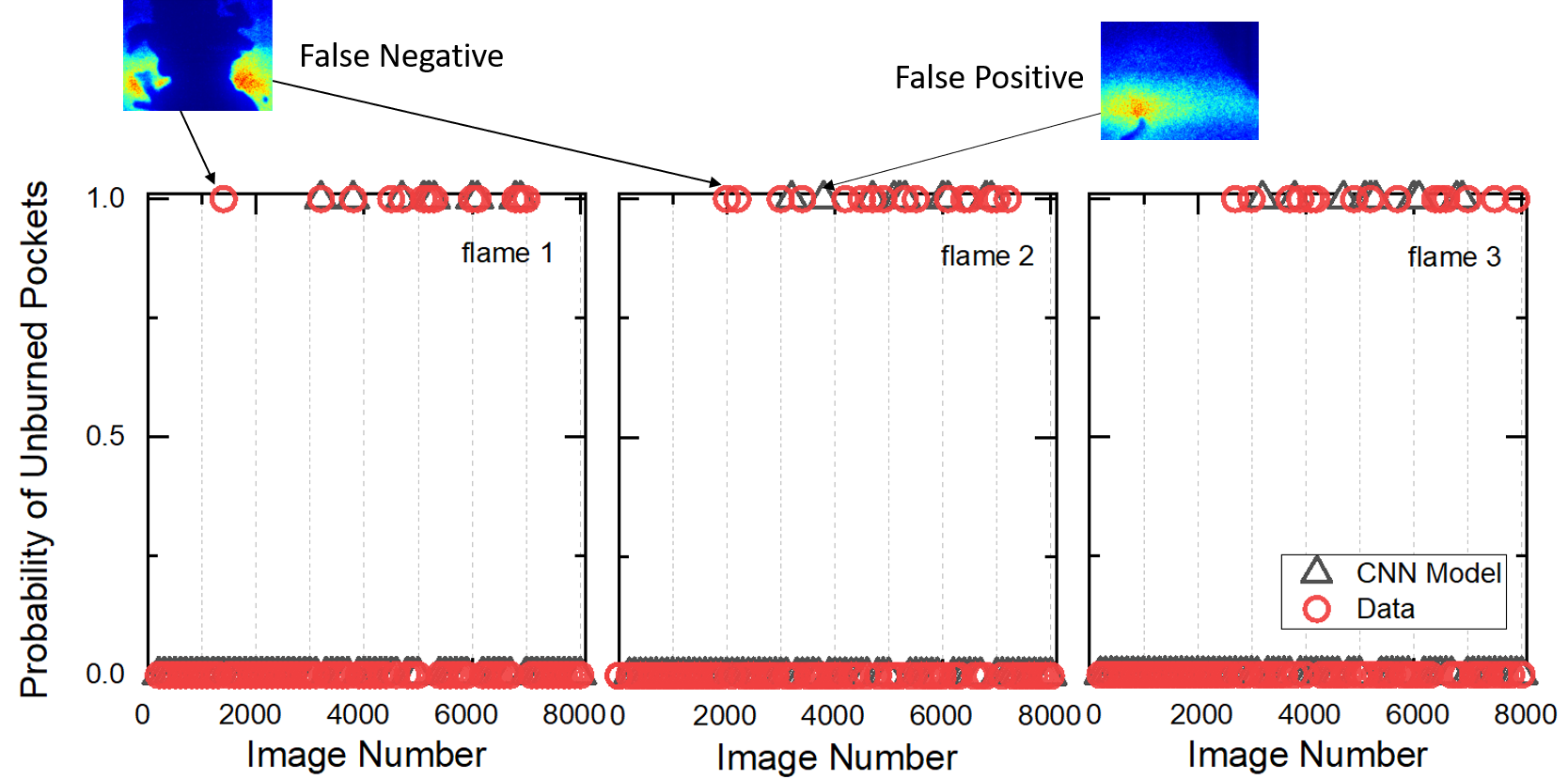}
\caption{Classification results for all flames. Each thousand images represent one panel of Fig. \ref{Fig:Flames}.}
\label{Fig:Results}
\end{figure}

Fig. \ref{Fig:Results} depicts the classification results of the CNN model for each flame. The probability of unburned pockets varies between zero and one as the image number increases. Each thousand images represent one panel of Fig. \ref{Fig:Flames}. The CNN model results and data are depicted with a hollow triangle and a hollow circle, respectively. Fig \ref{Fig:Results} and Table \ref{Table:Tests} clearly show that the classification problem is unbalanced. The number of pockets beyond the tip of the flame increases for flame 2 and 3.

\section{Conclusions}

Convolutional Neural Networks were used to classify high-speed OH-PLIF images in two different classes, images with one or more unburned pockets and images without an unburned pocket. The CNN model classified successfully 91.72\%, 89.35\% and 85.80\% of the images for flame 1, flame 2 and flame 3, respectively. The numbers of turbulent flame OH-PLIF images in the two classes were highly unbalanced thus challenging the methodologies normally used for balanced classes. CNN models effectively extract flame features relevant for unburned pocket detection under different CO$_2$ addition conditions. The present model provides a novel approach for OH-PLIF images classification.

\section{Acknowledgements}

The authors of Ref.\cite{Han2018} made the experimental data available for this study.

\bibliography{sample}

\end{document}